# On the Mirage of Long-Range Dependency, with an Application to Integer Multiplication

**Zichao Wei**
Saarland University
ziwe00001@stud.uni-saarland.de

## Abstract

Integer multiplication has long been considered a hard problem for neural networks, with the difficulty widely attributed to the $O(n)$ long-range dependency induced by carry chains. We argue that this diagnosis is wrong: long-range dependency is not an intrinsic property of multiplication, but a *mirage* produced by the choice of computational spacetime. We formalize the notion of mirage and provide a constructive proof: when two $n$-bit binary integers are laid out as a 2D outer-product grid, every step of long multiplication collapses into a $3 \times 3$ local neighborhood operation. Under this representation, a neural cellular automaton with only 321 learnable parameters achieves perfect length generalization up to $683 \times$ the training range. Five alternative architectures — including Transformer (6,625 params), Transformer+RoPE, and Mamba — all fail under the same representation. We further analyze how partial successes locked the community into an incorrect diagnosis, and argue that any task diagnosed as requiring long-range dependency should first be examined for whether the dependency is intrinsic to the task or induced by the computational spacetime.

## 1 Introduction

Integer multiplication is a notoriously difficult problem for neural networks — an open challenge that the community has repeatedly attempted but never fully resolved [1], [2], [3]. Standard Transformers [4] achieve near-zero length generalization on multiplication, and simply scaling up parameters does not help [1]. Multiplication has long been regarded as a litmus test for the fundamental reasoning capabilities of neural networks.

What draws our attention, however, is not only the remarkable ingenuity of recent work within the existing framework, but also the shared diagnosis of *why* multiplication is hard: it is a long-range dependency (LRD) problem, and therefore inherently difficult for Transformers.

But there is a paradox. The core of the Transformer is global attention [4] — a mechanism designed precisely to overcome the long-range dependency limitations of RNNs. That a mechanism *built for* LRD should fail on a task *because of* LRD [5], [6] — does this not strike one as absurd?

But should multiplication really be this hard? Consider long multiplication as taught in primary school: each step involves only two digits and a carry — the entire procedure is purely local. The long-range dependency that the community treats as the central difficulty simply does not exist in this process.

We argue that long-range dependency is not an intrinsic property of multiplication, but an artifact of the choice of *computational spacetime* — a mirage. We provide a constructive proof: a neural cellular automaton with only 321 parameters perfectly learns the local rules of multiplication and achieves $683 \times$ length generalization at 100% accuracy. Five alternative architectures — including

Transformer, RoPE, and Mamba — all fail under the same representation. §2 formalizes the notions of computational spacetime and mirage, §3 constructs the matching spacetime for multiplication, and §4 presents experimental validation.

## 2 Theoretical Framework

### 2.1 Computational Spacetime

What does the literature actually mean when it says "multiplication requires long-range dependency"?

Consider the most straightforward setup. Concatenate two $n$-bit binary integers $a$ and $b$ into a bit sequence and ask a model to output each bit of the product $c$. The $k$-th output bit $c_k$ depends on all partial products $\left(a_i, b_j\right)$ satisfying $i + j \leq k$, plus a carry chain that accumulates from low bits to high bits. This carry chain has length $O(L)$ — it spans the entire sequence. In this sense, multiplication does "require long-range dependency": each output bit can be influenced by distant input positions.

But there is an overlooked premise: the above description implicitly assumes two specific choices — encoding the input as a 1D sequence (a representation choice) and processing it with a model that accesses all positions in a single step (an architecture choice). Under a different encoding or a different computational model, the same task may exhibit an entirely different dependency structure.

To make this observation precise, we introduce the following concept.

**Definition 1 (Computational Spacetime).** A *computational spacetime* $G = (R, A)$ for task $f$ is jointly determined by two choices:

- *Representation* $R$ (spatial geometry): the encoding of data, which determines which information units are spatially "adjacent." A 1D sequence imposes a chain topology; a 2D grid imposes a lattice topology.
- *Architecture* $A$ (causal structure and time depth): the spatial range accessible at each computational step (causal connectivity), and the total number of steps permitted (time depth budget). A Transformer has global connectivity but fixed depth $O(1)$ [7], [8]; a cellular automaton has only local connectivity but can iterate for $O(n)$ steps.

Crucially, the choices of $R$ and $A$ constrain each other. Fixing a spatial geometry limits the viable causal structures — for instance, once data is arranged on a 2D grid, local iteration becomes the natural mode of computation, while global single-step processing becomes ill-suited. Conversely, fixing an architecture constrains the natural representation — Transformer's global attention has a natural affinity for 1D sequences, not 2D grids. Computational spacetime is an *indivisible whole*, not two independently adjustable knobs.

The central corollary of this definition is: all observable computational properties of task $f$ — including whether it exhibits long-range dependency — are *joint* properties of $(f, G)$, not of $f$ alone. Under $G_1 =$ (1D sequence, Transformer), multiplication exhibits $O(L)$ long-range dependency; but this does not mean multiplication *requires* long-range dependency — it only means that multiplication *under* $G_1$ does.[1]

### 2.2 Mirage

If long-range dependency is not an inherent property of the task but a joint product of $(f, G)$, a natural question arises: what happens to these dependencies when we switch to a different computational spacetime $G_2$?

[1]Compressing global information into a single symbol would require an exponentially large alphabet, violating the fixed-alphabet assumption of standard computational models. This transfers, rather than eliminates, the dependency.

**Definition 2 (Mirage).** A property $P$ (such as long-range dependency) is observed under computational spacetime $G_1$. If there exists another computational spacetime $G_2$ such that:

1. $P$ *vanishes* under $G_2$ — this is the necessary premise: if $P$ persists under all computational spacetimes, it is genuinely an intrinsic property of $f$ and cannot be a mirage;
2. $f$ under $G_2$ admits a *simpler rule description* (fewer learnable parameters, following the minimum description length principle) and *stronger generalization* (correctness beyond the training distribution) — this is the confirmatory evidence: the disappearance of $P$ is accompanied by an overall improvement in implementation, indicating that $G_2$ is closer to the task's causal structure.

Then $P$ is a *mirage* of $(f, G_1)$ — it is not an intrinsic property of $f$, but an artifact of $G_1$. We choose rule simplicity and generalization as confirmatory criteria because they are the two measures closest to causal mechanism discovery and least dependent on specific architectural details.[2]

### 2.3 Existence and the Core Question

Turing completeness guarantees that any computable function can be realized by purely local systems — Turing machines and cellular automata [9], [10], [11], [12] each depend only on finite neighborhoods per step. Long-range dependency is therefore always *eliminable*.[3] But eliminability does not imply mirage: a mirage requires that elimination yields simpler rules and stronger generalization. This paper provides a constructive answer for integer multiplication.

## 3 Construction

Let us return to the most elementary solution to multiplication: long multiplication. Regardless of the base, the algorithm consists of three local operations:

1. Compute partial products bit by bit
2. Align by weight position
3. Propagate carries bit by bit

We implement this algorithm in base-2, where each operation reduces to a binary function within a $3 \times 3$ neighborhood.

### 3.1 2D Outer-Product Encoding

Given two $n$-bit binary integers $a = (a_0, a_1, ..., a_{n-1})$ and $b = (b_0, b_1, ..., b_{n-1})$ (least significant bit first), we construct a $2n \times n$ grid $G$:

$$G_{i,j} = a_i \cdot b_j, \quad i \in \{0, ..., n-1\}, \quad j \in \{0, ..., n-1\} \tag{1}$$

The lower half $G_{i,j} = 0$ $(i \geq n)$ is reserved for carry extension. This outer-product grid encodes all partial products of the multiplication — $a_i \cdot b_j$ is exactly a contribution to bit position $i + j$.

### 3.2 Local Rules

The grid evolves under a uniform local rule $f$ that gathers partial products into column 0 and progressively resolves carries. Each step $G^{t+1} = f(G^t)$ consists of two parallel operations:

**Diagonal flow.** For all cells with $j > 0$, values move diagonally: $G^{t+1}_{i+1,j-1} += G^t_{i,j}$. This corresponds to the "align by weight" step of long multiplication — the partial product at $(i, j)$ contributes to bit position $i + j$, and reaches row $i + j$ of column 0 after $j$ diagonal steps.

**Carry.** For column 0, *before* receiving diagonal flow input, the current value undergoes binary carry resolution:

[2] Condition 1 alone is insufficient. The Game of Life can implement multiplication with LRD nominally "gone," but the encoding cost is astronomical and generalization is impossible. Condition 2 excludes such "elimination with degradation" cases.

[3] Iterative propagation — information transmitted cell by cell over multiple steps — does not constitute long-range dependency, just as light taking eight minutes to reach Earth from the Sun is not action at a distance.

$$\text{remainder}_i = G^t_{i,0} \bmod 2, \quad \text{carry}_i = G^t_{i,0} \div 2 \tag{2}$$

Column 0 is updated as: $G^{t+1}_{i,0} = \text{remainder}_i + \text{carry}_{i-1} + \text{flow input from } (i-1,1)$.

The key property: diagonal flow accesses only $(i,j)$ and $(i+1,j-1)$; carry accesses only $(i,0)$ and $(i-1,0)$. *All dependencies lie within a* $3 \times 3$ *neighborhood.*

After $O(n)$ steps, all partial products have flowed into column 0 and all carries have propagated. Column 0 then contains the binary representation of the product.

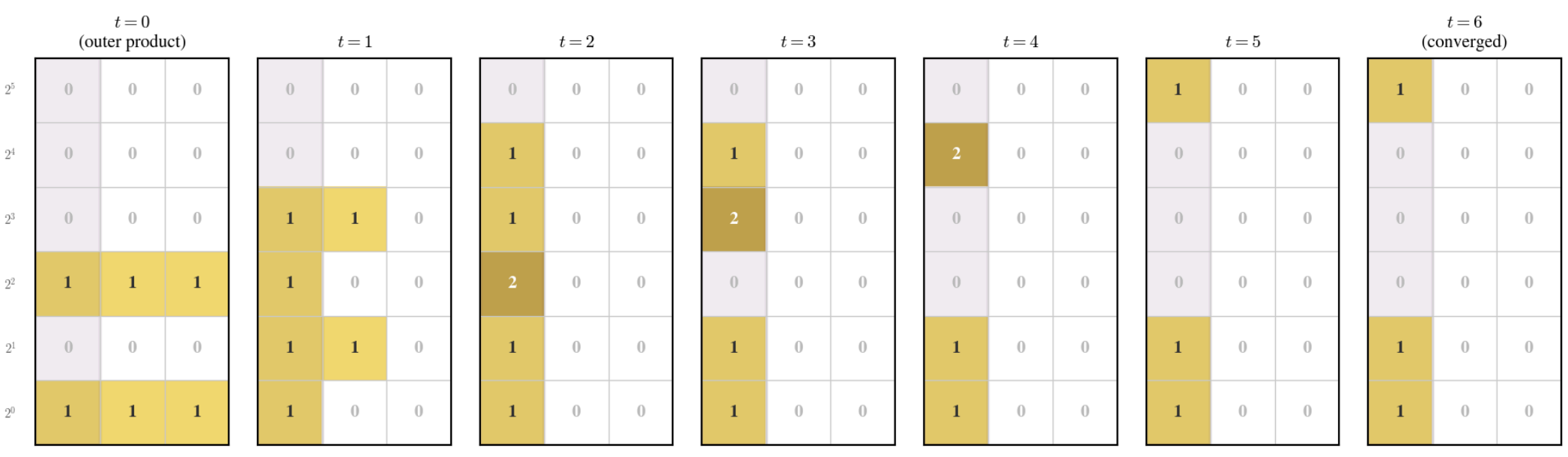

Figure 1: NCA evolution on the 2D outer-product grid: $5 \times 7 = 35$ (3-bit binary, 6 steps to convergence). At $t = 0$, the grid contains the outer product ($G_{i,j} = a_i \cdot b_j$). Diagonal flow gathers partial products into column 0 (highlighted in red); carries propagate locally via $\bmod 2$ / $\div 2$. All operations are within the $3 \times 3$ neighborhood.

### 3.3 Neural Cellular Automaton

The local rules above could be hand-coded as a CA program, but we take a more powerful approach: *learning* them with a neural network — completing the construction of computational spacetime $G_2$: $R$ = 2D outer-product encoding (spatial geometry), $A$ = $3 \times 3$ convolution + $O(n)$ iteration (causal structure).

A Neural Cellular Automaton (NCA) [13], [14] is a natively neuro-symbolic system [15]. Its *neural* side — a weight-sharing convolutional network — learns local rules from data via gradient descent; its *symbolic* side — the iterative dynamics of a cellular automaton — executes computation by repeatedly applying these rules. The key distinction is that computation arises not from a single feedforward pass, but from the *physical propagation* of local rules across the grid. The network provides the rule; evolution performs the reasoning.

Concretely, each inference step follows the *Relax-and-Project* mechanism of [15]:

$$G^{t+1} = \sigma(G^t + f_\theta(G^t)) \tag{3}$$

where $f_\theta$ is a two-layer convolutional network: $\text{Conv2d}(2 \to 16, k = 3) + \text{ReLU} + \text{Conv2d}(16 \to 1, k = 1)$, with 321 learnable parameters in total. The *relaxation* phase computes the update $f_\theta$ in continuous space, enabling gradient-based training. The *projection* phase $\sigma = \text{round}() + \text{clamp}(\min = 0)$ collapses the continuous output back to discrete integers — acting as a denoiser that zeroes out any numerical error below the decision boundary, ensuring lossless signal propagation over arbitrarily many iterations.

**Encoding.** Each cell's scalar value $v$ is expanded into a two-channel input $\{v, \cos(\pi v)\}$. The $\cos(\pi v)$ channel maps integer parity to $\{+1, -1\}$, making the $\bmod 2$ operation approximately linear for the network — a single linear combination suffices to implement carry detection. Grid boundaries are padded with $-1$ as a physical boundary condition (not a positional encoding), preserving translation invariance.

### 3.4 Chaos Training

Rather than training on complete multiplication trajectories, we adopt **Chaos Training**: learning the single-step local rule on random intermediate states.

Specifically, for each training sample: (1) randomly generate a pair of $n$-bit integers ($n \in \{2, 3, 4, 5, 6\}$) and construct the outer-product grid; (2) evolve the grid for $k$ steps using the ground-truth rule ($k$ sampled from $\mathrm{Uniform}(0, 4n)$), obtaining a random intermediate state $G^t$; (3) compute the next step $G^{t+1}$ using the ground-truth rule as the supervision signal.

The training loss is $\mathcal{L} = \mathrm{MSE}(f_\theta(G^t) + G^t, G^{t+1})$. This forces the model to learn the correct local rule across *all* possible intermediate states — not just along "correct trajectories." Chaos Training is critical: it ensures the model learns *the rule itself*, rather than memorizing specific trajectories.

## 4 Experiments

### 4.1 Setup

We evaluate the proposed method on binary integer multiplication. Training uses random pairs of 2- to 6-bit integers, Chaos Training for 30,000 steps (convergence at  4,000 steps), batch size 256, Adam optimizer, learning rate $10^{-3}$ with cosine annealing. At test time, for each bit length $n$, we sample random integer pairs (200 pairs for $n \leq 16$, 50 for $n > 16$), run iterative inference until a fixed point is reached, and compare against Python's arbitrary-precision arithmetic. We report exact-match accuracy. All experiments run on a single GPU.

### 4.2 Training and Generalization

**Training efficiency.** The NCA reaches 100% training accuracy (exact single-step prediction) at approximately 4,000 steps, with a total training time of  15 seconds. The model has only 321 learnable parameters — 1–3 orders of magnitude smaller than Transformer models on the same task. Training converges stably across 10 different random seeds (see Appendix).

**Length generalization.** Table 1 reports key milestones. The model achieves 100% accuracy within the training range (2–6 bits) and maintains 100% on all tested out-of-distribution lengths — from 16 bits up to 4,096 bits (2,467 decimal digits), achieving **683 × length generalization**. The number of inference steps scales strictly linearly with input length ($\approx n + 2$ steps), matching the $O(n)$ theoretical lower bound of long multiplication. Full results are in Appendix Table 2.

**Mirage diagnosis.** We can now verify the mirage conditions from §2. Condition 1 (necessary premise): under the computational spacetime $G_2$ of 2D outer-product encoding + NCA, all operations are confined to $3 \times 3$ neighborhoods — long-range dependency vanishes. Condition 2 (confirmatory evidence): 321 learnable parameters — the rule description is 1–3 orders of magnitude simpler than existing methods; 683 $\times$ perfect length generalization — generalization is two orders of magnitude stronger than the previous best. Both conditions are satisfied: **the $O(L)$ long-range dependency of multiplication under the 1D Transformer spacetime is confirmed as a mirage.**

Table 1: NCA length generalization (321 parameters, trained on 2–6 bits). 100% exact-match accuracy at all tested lengths.

| Bits | Split | Accuracy | NCA Steps | Gen. Factor |
|---|---|---|---|---|
| 2-6 | Train | 100% | 1-6 | 1× |
| 16 | OOD | 100% | 18 | 2.7× |
| 64 | OOD | 100% | 67 | 10.7× |
| 256 | OOD | 100% | 262 | 42.7× |
| 1024 | OOD | 100% | 1,032 | 170.7× |
| 4096 | OOD | 100% | 4,106 | 682.7× |

### 4.3 Alternative Architectures

A natural question is: can other architectures exploit the same 2D outer-product encoding without NCA? We tested five alternatives — Transformer (with sinusoidal PE and RoPE [16] variants), Mamba [17], and MLP — all sharing identical input encoding, training protocol, and inference mechanism. As shown in Figure 2, all five fail completely. Notably, every model — including Transformer (6,625 params) and Mamba (4,577 params) — achieves 100% single-step prediction accuracy during Chaos Training. They all fit the single-step training target perfectly. Failure occurs only during multi-step iterative inference.

The problem is not that these architectures cannot fit the single-step target, but that they *cannot stably iterate* the learned update. Transformer's softmax normalization makes the attention distribution size-dependent — distributions learned on training-sized grids become noise on larger test grids. Adding positional encodings (sinusoidal or RoPE [16], [18]) not only fails to help but introduces additional size-dependent spurious patterns. Mamba [17] flattens the 2D grid into a 1D sequence for scanning [19], destroying spatial adjacency and likewise failing to maintain stable iteration.

This is not an ablation — computational spacetime is an indivisible whole, and changing the architecture changes the entire spacetime. This is a comparison of six different computational spacetimes, of which only the locally matched one succeeds.

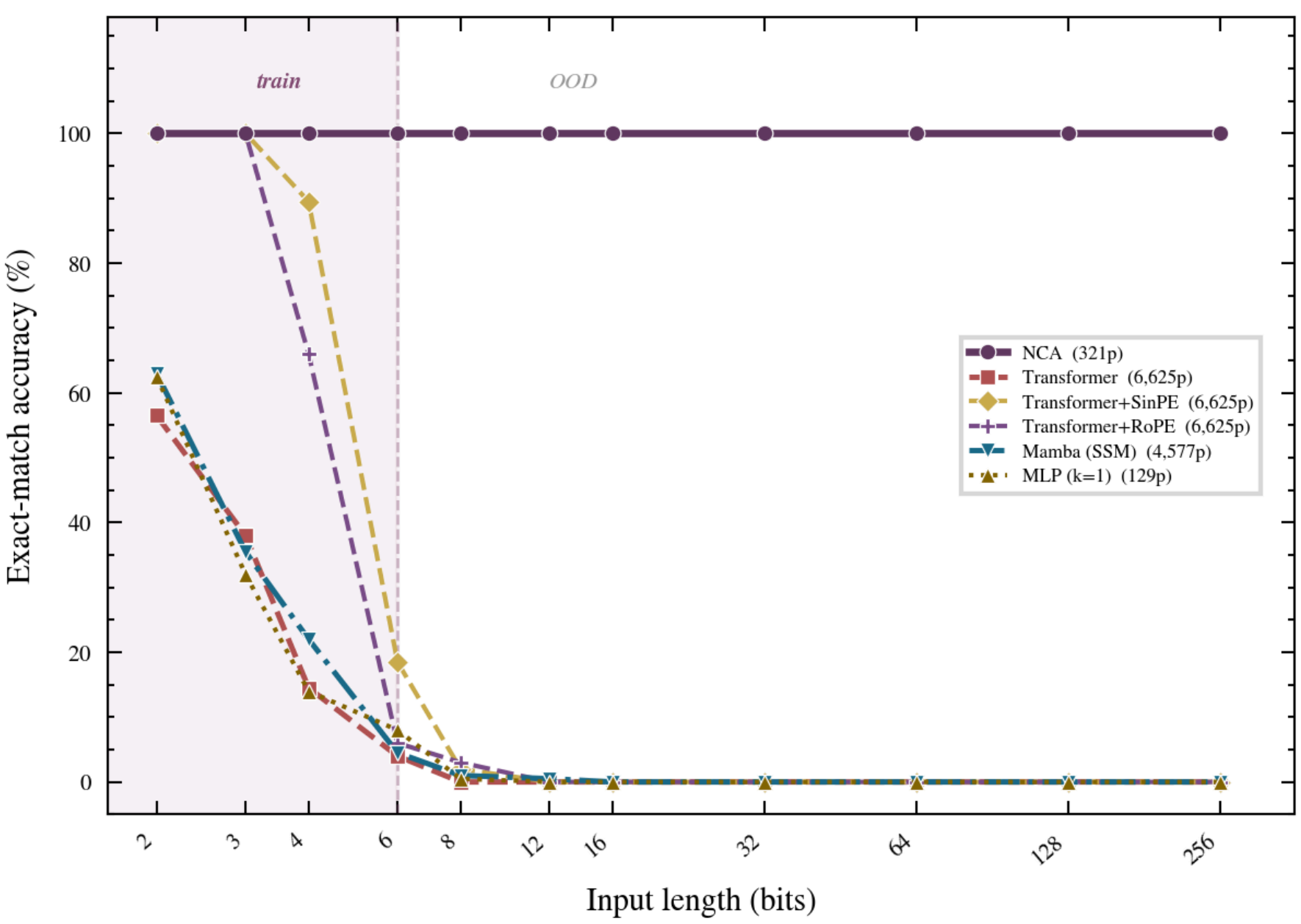

Figure 2: Length generalization curves for six architectures under the same 2D outer-product encoding. NCA (321 params) maintains 100% accuracy at all test lengths; the other five architectures (including Transformer, Transformer+RoPE, Mamba) collapse to 0% beyond the training distribution.

## 5 Related Work

Research on neural network multiplication can be unified through the lens of computational spacetime. The path the community has taken is, in essence, a history of incremental patching within mismatched spacetimes.

**Pure string transduction.** The most basic setup treats multiplication as a digit-string-to-digit-string mapping, training Transformers end-to-end. Under this setup, both dimensions of the computational spacetime $G$ are mismatched: the spatial geometry is a 1D sequence (flattening the 2D partial products), and the causal structure is global attention with fixed $O(1)$ depth. The baseline experiments of L. Qiu, J. Li, C. Su, C. J. Zhang, and L. Chen [1] and X. Bai *et al.* [2] fall into this category [20], [21], [22], showing that multiplication is extremely difficult to learn under this $G$. Yet these experiments also demonstrate an important converse: at *fixed lengths*, Transformers have sufficient capacity to fit multiplication [3], [23] — the bottleneck is not model capacity itself, but the choice of computational spacetime.

**In-framework patching.** The second line of work accepts the existing spacetime and attempts to survive within $G$ through training tricks: reversing output order to align carry direction, increasing depth, curriculum learning from short to long, and balanced sampling to reduce data bias. The strongest result of L. Qiu, J. Li, C. Su, C. J. Zhang, and L. Chen [1] — improving $5 \times 5$ decimal multiplication from 0% to 99.9% — was achieved through careful combination of such techniques. But these improvements operate only at *fixed lengths* and do not address the root cause: the computational spacetime itself has not changed.

**Partial spacetime correction.** The third line of work recognizes that some dimension of $G$ must change. Chain-of-thought [24], [25], [26] and scratchpad methods allow models to generate intermediate computation steps, *restoring time depth* within a fixed 1D spatial geometry — each generated token reclaims one computational step. This partially corrects the temporal mismatch of $G$, yielding some improvement in generalization. But the spatial dimension remains 1D with global attention at each step, leaving the spatial mismatch untouched — hence the improvement is limited. X. Bai *et al.* [2]'s implicit chain-of-thought (ICoT) and auxiliary losses push further in this direction [27]. Position coupling H. Cho, J. Cha, S. Bhojanapalli, and C. Yun [28] and priming S. Jelassi, S. d'Ascoli, C. Domingo-Enrich, Y. Wu, Y. Li, and F. Charton [29] approach from another dimension, attempting to simulate 2D positional alignment at the attention level — partially correcting the spatial geometry. This line has achieved substantive progress — generalization improved from zero to 2–3× [30], [31] — but remains a local patch to $G$, not a wholesale replacement. Other work attempts to restore time depth through recurrent architectures [32], [33], [34], but the spatial geometry remains unchanged.

These three stages form a clear progression, unified by the degree of spacetime matching: pure transduction ($G$ fully mismatched) $\rightarrow$ in-framework patching (no change to $G$) $\rightarrow$ partial correction (fixing one dimension of $G$). Each step brings improvement, but each improvement also reinforces an implicit assumption: long-range dependency is the real difficulty, and the direction of patching is correct. A deeper analysis of this self-locking effect follows in §6.

## 6 Discussion

### 6.1 Simplicity Beneath the Mirage

The experimental results of this paper — 321 parameters, $683 \times$ length generalization, 100% exact match — appear surprising at first glance. But the correct interpretation is not "we are better at teaching machines to multiply" but rather "multiplication was never as hard as people thought."

Long multiplication is an algorithm taught in primary school: multiply digit by digit, stack diagonally, add columns, propagate carries — each step involves only adjacent cells. The reason 321 parameters suffice is not that our model is clever, but that these local rules are inherently simple. Previous work required thousands or millions of parameters not because multiplication demands them, but because their computational spacetime turned a simple problem into a hard one.

Yet a mirage is not an illusion — it is genuinely observable. Just as an optical mirage is a real optical phenomenon (light truly refracts in a temperature gradient), carry chains under $G_1$ genuinely manifest as $O(n)$ long-range dependency — real and actionable, but a property of $G_1$, not of multiplication. Upon switching to $G_2$, the dependency is not "resolved" — it is revealed to have never belonged to the task in the first place.

### 6.2 The Question Was Asked Backwards

If the mirage is so easy to dispel, why did the community take so long to find it? Because the question was asked backwards. The community studied multiplication not to understand multiplication, but to probe the limits of Transformers. The question was *why can't Transformers do* *multiplication?* — with the Transformer as subject, answers were sought within its framework: attention is insufficient, depth is insufficient, positional encoding is wrong.

This paper asks a different question: *what does multiplication itself require?* Starting from the task, one first sees the local causal structure of long multiplication, then the representation emerges naturally (2D outer product), then the architecture is constrained (local iteration), and 321 parameters suffice. There is a causal asymmetry at work. The correct direction runs from the task's causal structure to the representation that respects it, and only then to the architecture that implements it. The community proceeded in reverse: starting from the Transformer, they adopted its natural representation — the 1D sequence — which manufactured the mirage of long-range dependency. Reasoning from the architecture outward, one is already standing inside the spacetime that produces the mirage.

The three stages in §5 are the unfolding of this self-locking. CoT restores time depth without changing the representation — partially effective. Position coupling corrects spatial alignment without changing the architecture — also partially effective. Each fixes half the problem; both reinforce the belief that LRD is the real difficulty. Had the patches failed entirely, the premise might have been questioned; precisely because they *partially* work, the incorrect diagnosis becomes entrenched. Conversely, NCA's locality constraint *forces* the question: how should data be arranged so that operations fall within a $3 \times 3$ neighborhood? The convenience of action at a distance extinguishes this question; locality gives rise to the correct representation.

### 6.3 Limitations

Our experiments are limited to binary multiplication. Base-2 strips away the numerical complexity of base-10 and exposes the skeletal structure of multiplication — which is local. Decimal carry rules involve a larger state space ($0 - 81$ rather than $0 - 1$) and may require more elaborate encoding, but this is an increase in encoding complexity, not a loss of locality. Another limitation is computational cost: NCA iterates $O(n)$ steps on $n^2$ cells, yielding $O(n^3)$ total FLOPs, higher than the Transformer's $O(n^2)$. But our claim is not "our method is faster" — it is "long-range dependency in multiplication is a mirage," an epistemological proposition about the nature of the problem, not a competition in engineering efficiency. Likewise, we do not claim that all long-range dependencies in all tasks are mirages, nor that NCA is a universal paradigm. What we do claim is a principle: before attributing difficulty to the task itself, first examine whether your computational spacetime manufactured that difficulty.

## 7 Conclusion

Through the case of integer multiplication, this paper argues a general proposition: widely accepted long-range dependency may be a mirage manufactured by the choice of computational spacetime, rather than an intrinsic property of the task. When multiplication is placed in a matching computational spacetime, 321 parameters suffice to encode the complete, arbitrarily generalizable computational rule. This result should not be read as our having designed a better multiplication model, but as multiplication never having been as hard as believed — the difficulty was manufactured by the spacetime.

But the true value of multiplication lies beyond multiplication itself. Multiplication became a research object not because the world needs a neural network to replace hardware multipliers, but because it exposes a systematic deficiency in the current paradigm's capacity for exact reasoning. The value of multiplication lies precisely in its *catastrophic* nature: the Transformer's total failure on multiplication is too conspicuous to ignore, forcing the community to confront the problem head-on. But what if computational spacetime mismatch does not always cause catastrophic failure? What if, on more consequential tasks — reasoning, planning, scientific discovery — mirages produce not a collapse from 100% to 0%, but a quiet degradation from 95% to 80%? Such degradation would not make headlines, would not trigger dedicated benchmarks; it would be filed under "limitations of current methods" rather than "fundamental paradigm deficiency." Multiplication is the canary — its death is unambiguous and dramatic, effectively warning of danger in the mine. But the real concern is not the canary; it is the as-yet-uninspected tunnels where the air appears normal.

The mirage of long-range dependency is real, measurable, and even partially improvable — but it is not how things truly are. The next time you observe long-range dependency in a task, consider switching computational spacetimes first — and see whether it is still there.

## References

[1] L. Qiu, J. Li, C. Su, C. J. Zhang, and L. Chen, "Dissecting Multiplication in Transformers: Insights into LLMs," no. arXiv:2407.15360. arXiv, July 2024. doi: 10.48550/arXiv.2407.15360.

[2] X. Bai *et al.*, "Why Can't Transformers Learn Multiplication? Reverse-engineering Reveals Long-Range Dependency Pitfalls," no. arXiv:2510.00184. arXiv, Sept. 2025. doi: 10.48550/arXiv.2510.00184.

[3] A. Gambardella, Y. Iwasawa, and Y. Matsuo, "Language Models Do Hard Arithmetic Tasks Easily and Hardly Do Easy Arithmetic Tasks," no. arXiv:2406.02356. arXiv, June 2024. doi: 10.48550/arXiv.2406.02356.

[4] A. Vaswani *et al.*, "Attention Is All You Need," in *Advances in Neural Information Processing Systems*, Curran Associates, Inc., 2017.

[5] M. Hahn, "Theoretical Limitations of Self-Attention in Neural Sequence Models," *Transactions of the Association for Computational Linguistics*, vol. 8, pp. 156–171, Jan. 2020, doi: 10.1162/tacl_a_00306.

[6] M. Hahn and M. Rofin, "Why Are Sensitive Functions Hard for Transformers?," no. arXiv:2402.09963. arXiv, May 2024. doi: 10.48550/arXiv.2402.09963.

[7] S. Bhattamishra, K. Ahuja, and N. Goyal, "On the Ability and Limitations of Transformers to Recognize Formal Languages," no. arXiv:2009.11264. arXiv, Oct. 2020. doi: 10.48550/arXiv.2009.11264.

[8] W. Merrill and A. Sabharwal, "The Parallelism Tradeoff: Limitations of Log-Precision Transformers," *Transactions of the Association for Computational Linguistics*, vol. 11, pp. 531–545, June 2023, doi: 10.1162/tacl_a_00562.

[9] M. Cook, "Universality in Elementary Cellular Automata," *Complex Systems*, vol. 15, no. 1, pp. 1–40, Mar. 2004, doi: 10.25088/ComplexSystems.15.1.1.

[10] S. Wolfram, "Computation Theory of Cellular Automata," *Communications in Mathematical Physics*, vol. 96, no. 1, pp. 15–57, Mar. 1984, doi: 10.1007/BF01217347.

[11] S. Wolfram, *A New Kind of Science*. Champaign (Ill.): Wolfram, 2002.

[12] J. Von Neumann and A. W. (. W. Burks, *Theory of Self-Reproducing Automata*. Urbana, University of Illinois Press, 1966.

[13] A. Mordvintsev, E. Randazzo, E. Niklasson, and M. Levin, "Growing Neural Cellular Automata," *Distill*, vol. 5, no. 2, p. e23, Feb. 2020, doi: 10.23915/distill.00023.

[14] W. Gilpin, "Cellular Automata as Convolutional Neural Networks," *Physical Review E*, vol. 100, no. 3, p. 32402, Sept. 2019, doi: 10.1103/PhysRevE.100.032402.

[15] Z. Wei, "On the Spatiotemporal Dynamics of Generalization in Neural Networks," no. arXiv:2602.01651. arXiv, Feb. 2026. doi: 10.48550/arXiv.2602.01651.

[16] J. Su, M. Ahmed, Y. Lu, S. Pan, W. Bo, and Y. Liu, "RoFormer: Enhanced Transformer with Rotary Position Embedding," *Neurocomputing*, vol. 568, p. 127063, Feb. 2024, doi: 10.1016/j.neucom.2023.127063.

[17] A. Gu and T. Dao, "Mamba: Linear-time Sequence Modeling with Selective State Spaces," in *First Conference on Language Modeling*, Aug. 2024.

[18] L. Zhao *et al.*, "Length Extrapolation of Transformers: A Survey from the Perspective of Positional Encoding," no. arXiv:2312.17044. arXiv, Apr. 2024. doi: 10.48550/arXiv.2312.17044.

[19] J. Park *et al.*, "Can Mamba Learn How to Learn? A Comparative Study on in-Context Learning Tasks," no. arXiv:2402.04248. arXiv, Apr. 2024. doi: 10.48550/arXiv.2402.04248.

[20] T. Tuncer *et al.*, "Solving the Multiplication Problem of a Large Language Model System Using a Graph-Based Method," no. arXiv:2310.13016. arXiv, Oct. 2023. doi: 10.48550/arXiv.2310.13016.

[21] A. B. de Luca, G. Giapitzakis, and K. Fountoulakis, "Learning to Add, Multiply, and Execute Algorithmic Instructions Exactly with Neural Networks," no. arXiv:2502.16763. arXiv, Jan. 2026. doi: 10.48550/arXiv.2502.16763.

[22] G. Delétang *et al.*, "Neural Networks and the Chomsky Hierarchy," no. arXiv:2207.02098. arXiv, Feb. 2023. doi: 10.48550/arXiv.2207.02098.

[23] H. Jiang, M. Hahn, G. Zetzsche, and A. W. Lin, "Softmax Transformers Are Turing-Complete," no. arXiv:2511.20038. arXiv, Nov. 2025. doi: 10.48550/arXiv.2511.20038.

[24] J. Wei *et al.*, "Chain-of-Thought Prompting Elicits Reasoning in Large Language Models," no. arXiv:2201.11903. arXiv, Jan. 2023. doi: 10.48550/arXiv.2201.11903.

[25] M. Nye *et al.*, "Show Your Work: Scratchpads for Intermediate Computation with Language Models," no. arXiv:2112.00114. arXiv, Nov. 2021. doi: 10.48550/arXiv.2112.00114.

[26] G. Feng, B. Zhang, Y. Gu, H. Ye, D. He, and L. Wang, "Towards Revealing the Mystery behind Chain of Thought: A Theoretical Perspective," *Advances in Neural Information Processing Systems*, vol. 36, pp. 70757–70798, Dec. 2023.

[27] C. Li *et al.*, "Chain of Code: Reasoning with a Language Model-Augmented Code Emulator," no. arXiv:2312.04474. arXiv, July 2024. doi: 10.48550/arXiv.2312.04474.

[28] H. Cho, J. Cha, S. Bhojanapalli, and C. Yun, "Arithmetic Transformers Can Length-Generalize in Both Operand Length and Count," no. arXiv:2410.15787. arXiv, Apr. 2025. doi: 10.48550/arXiv.2410.15787.

[29] S. Jelassi, S. d'Ascoli, C. Domingo-Enrich, Y. Wu, Y. Li, and F. Charton, "Length Generalization in Arithmetic Transformers," no. arXiv:2306.15400. arXiv, June 2023. doi: 10.48550/arXiv.2306.15400.

[30] H. Zhou *et al.*, "What Algorithms Can Transformers Learn? A Study in Length Generalization," no. arXiv:2310.16028. arXiv, Oct. 2023. doi: 10.48550/arXiv.2310.16028.

[31] X. Huang *et al.*, "A Formal Framework for Understanding Length Generalization in Transformers," no. arXiv:2410.02140. arXiv, Apr. 2025. doi: 10.48550/arXiv.2410.02140.

[32] M. Dehghani, S. Gouws, O. Vinyals, J. Uszkoreit, and Ł. Kaiser, "Universal Transformers," no. arXiv:1807.03819. arXiv, Mar. 2019. doi: 10.48550/arXiv.1807.03819.

[33] Y. Fan, Y. Du, K. Ramchandran, and K. Lee, "Looped Transformers for Length Generalization," no. arXiv:2409.15647. arXiv, May 2025. doi: 10.48550/arXiv.2409.15647.

[34] A. Graves, G. Wayne, and I. Danihelka, "Neural Turing Machines," no. arXiv:1410.5401. arXiv, Dec. 2014. doi: 10.48550/arXiv.1410.5401.

# A Appendix

## A.1 Full Generalization Results

Table 2: Full generalization results. NCA (321 parameters) achieves 100% exact-match accuracy at all tested lengths. Inference steps scale strictly linearly with bit length.

| Bits | Split | Accuracy | NCA Steps | Decimal Digits | Gen. Factor |
|---|---|---|---|---|---|
| 2-6 | Train | 100% | 1-6 | 1-4 | 1× |
| 8 | OOD | 100% | 8 | 5 | 1.3× |
| 16 | OOD | 100% | 18 | 10 | 2.7× |
| 32 | OOD | 100% | 35 | 20 | 5.3× |
| 64 | OOD | 100% | 67 | 39 | 10.7× |
| 128 | OOD | 100% | 133 | 78 | 21.3× |
| 256 | OOD | 100% | 262 | 155 | 42.7× |
| 512 | OOD | 100% | 519 | 309 | 85.3× |
| 1024 | OOD | 100% | 1,032 | 617 | 170.7× |
| 2048 | OOD | 100% | 2,056 | 1,234 | 341.3× |
| 4096 | OOD | 100% | 4,106 | 2,467 | 682.7× |

## A.2 Model Scale and Robustness

We test generalization success rate across 4 hidden dimensions × 10 random seeds to identify the minimum reliable configuration.

Table 3: Model scale and generalization robustness. 81 parameters is near the theoretical lower bound, with some seeds succeeding; 161+ parameters converge reliably to the correct rule.

| Hidden | Params | Success (10 seeds) | Rate | Failure Mode |
|---|---|---|---|---|
| 4 | 81 | 3 / 10 | 30% | carry misalign (5), gradual decay (2) |
| 8 | 161 | 9 / 10 | 90% | carry misalign (1) |
| 16 | 321 | 10 / 10 | 100% | — |
| 32 | 641 | 10 / 10 | 100% | — |

### A.3 Architecture Details for Alternative Models

All alternative architectures share the same input encoding ($\{v, \cos(\pi v)\}$ two-channel parity encoding), the same training protocol (Chaos Training, 10,000 steps, Adam with cosine annealing), and the same inference mechanism (residual update + $\mathrm{round}() + \mathrm{clamp}(\min = 0)$, iterated until fixed point).

Table 4: Architecture details for all tested models. All Transformer variants flatten the 2D grid to a 1D token sequence. The Mamba implementation uses the core selective state space mechanism (input-dependent $B$, $C$, $\Delta$ projections with linear recurrence) without the CUDA-optimized parallel scan of the official library; the failure mode — 1D flattening destroying 2D spatial adjacency — is independent of this implementation choice.

| Model | Params | Architecture | Locality |
|---|---|---|---|
| NCA | 321 | Conv2d(2$\rightarrow$16, $k$=3) + ReLU + Conv2d(16$\rightarrow$1, $k$ =1) | 3×3 local |
| Transformer | 6,625 | $d$=16, 4 heads, 2 layers, no PE | global |
| Trans.+SinPE | 6,625 | same + 2D sinusoidal PE | global + abs pos |
| Trans.+RoPE | 6,625 | same + 2D rotary PE applied to Q/K | global + rel pos |
| Mamba | 4,577 | bidirectional selective SSM, $d$=16, $d_{\text{state}}$=16, 2 layers | 1D sequential |
| MLP | 129 | Conv2d(2$\rightarrow$32, $k$=1) + ReLU + Conv2d(32$\rightarrow$1, $k$ =1) | pointwise |